%% file: main.tex
\documentclass[runningheads]{llncs}
\usepackage[T1]{fontenc}
\usepackage{graphicx}
\usepackage{subcaption}
\usepackage{amsmath}
\usepackage{booktabs}
\usepackage[table,xcdraw]{xcolor}
\usepackage[font=small]{caption}
\usepackage{algorithm}
\usepackage[noend]{algpseudocode}
\usepackage{amssymb} 
\usepackage{tabularx}
\usepackage{xspace}
\usepackage{array}
\usepackage{booktabs}
\usepackage{adjustbox,makecell}
\usepackage[inline]{enumitem}
\usepackage[pagebackref=false,breaklinks,colorlinks]{hyperref}

\usepackage[capitalize,nameinlink]{cleveref}
\crefformat{figure}{#2\figurename~(#1)#3}

\input{preamble}
\newenvironment{credits}{\section*{Credits}}{}

\begin{document}

\title{D-MASTER: Mask Annealed Transformer for Unsupervised Domain Adaptation in Breast Cancer Detection from Mammograms}

\titlerunning{\myname}

\author{Tajamul Ashraf\inst{1}\textsuperscript{\textdagger}\orcidID{0000-0002-7372-3782} \and
Krithika Rangarajan\inst{2}\textsuperscript{\textasteriskcentered}\orcidID{0000-0001-5376-6390} \and
Mohit Gambhir\inst{2}\textsuperscript{\textasteriskcentered}\orcidID{0000-0001-7071-0460} \and 
Richa Gabha\inst{2}\textsuperscript{\textasteriskcentered}\orcidID{0009-0007-1872-0670} \and
Chetan Arora\inst{1}\orcidID{0000-0003-0155-0250}}


\renewcommand{\thefootnote}{\textdagger}
\footnotetext{Corresponding author}
\renewcommand{\thefootnote}{\textasteriskcentered}
\footnotetext{Radiologists who annotated the \texttt{RSNA-BSD1K} dataset}

\authorrunning{Tajamul et al.}
%
\institute{Indian Institute of Technology Delhi, New Delhi 110016, INDIA\\
\email{tajamul@sit.iitd.ac.in}\and
AIIMS, Delhi, New Delhi 110029, INDIA \\
}

\maketitle             

\input{Sections/abstract}

\input{Sections/introduction}
\input{Sections/methodology}

\input{Sections/results}

\input{Sections/conclusion}

\input{Sections/acknowledgment}

\bibliographystyle{splncs04}
\bibliography{refs}

\clearpage
\setcounter{page}{1}
\renewcommand\thefigure{S\arabic{figure}}
\setcounter{figure}{0}
\renewcommand\thetable{S\arabic{table}}
\setcounter{table}{0}
\appendix

\input{Sections/supplementary}

\end{document}

%% file: preamble.tex
\def\dnn{\texttt{DNN}\xspace}
\def\dnns{\texttt{DNNs}\xspace}
\def\bcdm{\texttt{BCDM}\xspace}

\def\fpi{\texttt{FPI}\xspace}
\def\myname{\texttt{D-MASTER}\xspace}
\def\sota{\texttt{SOTA}\xspace}
\def\uda{\texttt{UDA}\xspace}
\def\inbreast{\texttt{INBreast}\xspace}
\def\ddsm{\texttt{DDSM}\xspace}
\def\rsna{\texttt{RSNA-BSD1K}\xspace}
\def\mae{\texttt{MAE}\xspace}
\def\maes{\texttt{MAEs}\xspace}
\def\D{\mathcal{D}\xspace}
\def\L{\mathcal{L}\xspace}

\def\defdetr{\texttt{DefDETR}\xspace}
\def\ema{\texttt{EMA}\xspace}
\def\froc{\texttt{FROC}\xspace}

\newcommand{\myfirstpara}[1]{\par \noindent \textbf{#1.}}
\newcommand{\mypara}[1]{\vspace{0.2em} \myfirstpara{#1}}

%% file: Sections/abstract.tex
\begin{abstract}

We focus on the problem of Unsupervised Domain Adaptation (\uda) for breast cancer detection from mammograms (\bcdm) problem. Recent advancements have shown that masked image modeling serves as a robust pretext task for \uda. However, when applied to cross-domain \bcdm, these techniques struggle with breast abnormalities such as masses, asymmetries, and micro-calcifications, in part due to the typically much smaller size of region of interest in comparison to natural images. This often results in more false positives per image (\fpi) and significant noise in pseudo-labels typically used to bootstrap such techniques. Recognizing these challenges, we introduce a transformer-based Domain-invariant Mask Annealed Student Teacher autoencoder (\myname) framework. \myname adaptively masks and reconstructs multi-scale feature maps, enhancing the model's ability to capture reliable target domain features. \myname also includes adaptive confidence refinement to filter pseudo-labels, ensuring only high-quality detections are considered. We also provide a bounding box annotated subset of 1000 mammograms from the RSNA Breast Screening Dataset (referred to as \rsna) to support further research in \bcdm. We evaluate \myname on multiple \bcdm datasets acquired from diverse domains. Experimental results show a significant improvement of 9\% and 13\% in sensitivity at 0.3 \fpi over state-of-the-art \uda techniques on publicly available benchmark \inbreast and \ddsm datasets respectively. We also report an improvement of 11\% and 17\% on In-house and \rsna datasets respectively. The source code, pre-trained \myname model, along with \rsna dataset annotations is available at \href{https://dmaster-iitd.github.io/webpage/}{d-master/webpage}.
%
\end{abstract}

%% file: Sections/introduction.tex
\section{Introduction}

Deep neural networks (\dnns) have achieved noteworthy breakthroughs in medical image analysis~\cite{fang2022deep,sun2022transformer,shu2020deep,cai2022advanced,xie2023deep,gonzalez2023robust} and have shown exceptional performance in specific tasks such as breast cancer detection on mammography. However, they suffer from relatively lower performance when there is a distribution gap between the training data and the deployed environment. This effect is particularly pronounce in medical imaging problems, due to relatively smaller size, and fewer number of annotated datasets, which does not allow a \dnn model to capture domain-invariant features. This affects the generalisability of the network across different geographies, different machines, techniques, and protocols of image acquisition. While images in a target population may be available to fine-tune a model, annotations are usually more expensive due to unavailability of a medical expert. Thus, in medical imaging problems, there is a strong need for effective methodologies in unsupervised domain adaptation (\uda).

\input{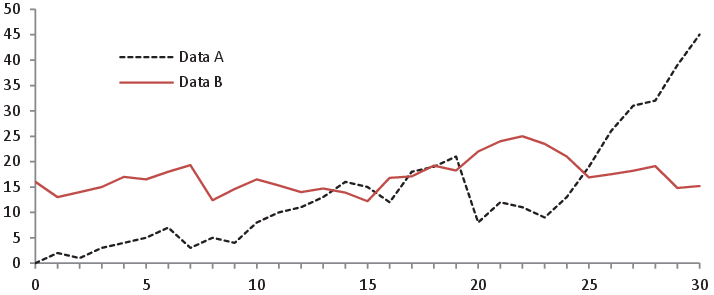} 

\uda has been extensively studied in case of natural images,
utilizing techniques such as adversarial learning~\cite{li2022cross},pseudo-label training~\cite{deng2021unbiased,li2022cross,yu2022cross}, image-to-image translation~\cite{yu2022cross}, graph reasoning~\cite{li2022sigma}, and adaptive mean Teacher training~\cite{deng2021unbiased}, improving domain adaptation efficiency of object detectors. 
Recently multiple works focused on using Mask autoencoders (\mae) methods in large-scale pretraining for vision models, involving masking parts of input and reconstructing them~\cite{he2022masked,tong2022videomae}. However, these approaches overlook domain shifts. Alternatively, widespread utilization of Teacher-Student models, wherein a Teacher provides pseudo-labels for target domain (unlabeled images) to supervise a Student model, leads to notable adaptation improvements~\cite{yu2022cross,zhao2023masked}. However, these techniques face the challenge of incorrect predictions and excessive false positives per image due to low-quality pseudo-labels, particularly for medical imaging problems. Pseudo-labels are filtered from the Teacher model's outputs based on the confidence score threshold. Selecting numerous pseudo-labels with low thresholds leads to inclusion of incorrect predictions, and compromising performance. Conversely, higher thresholds yield a limited number of pseudo-boxes, resulting in sub-optimal supervision. Existing Teacher-Student models often produce pseudo-labels riddled with errors and false positives, as illustrated in \cref{f1a} and \cref{f1b}. Though~\cite{li2022cross,yu2022cross,zhao2023masked} utilize techniques like adversarial alignment, weak-strong augmentation, and selective retraining of Student model to minimize the false positives in pseudo-labels, these approaches fail on medical images. 

\mypara{Contributions of this work}
We note that screening mammography inherently differs from natural images, with breast abnormalities such as masses, asymmetries, and micro-calcifications, typically much smaller in comparison to the salient objects present in natural images, emphasizing the need for approaches specific to this problem. To address these issues, we make following contributions in this work:
\begin{enumerate*}[label=\textbf{(\arabic*)}]
\item We introduce \myname, a transformer-based Domain-invariant Mask Annealed Student Teacher Autoencoder Framework for cross-domain breast cancer detection from mammograms (\bcdm), integrating a novel mask-annealing technique and adaptive confidence refinement module. Unlike pretraining with mask autoencoders (\maes)~\cite{he2022masked}, leveraging massive datasets for training and then fine-tuning on smaller datasets, we present a novel learnable masking technique for the \mae branch that generates masks of different complexities, which are reconstructed by the \defdetr~\cite{zhu2020deformable} encoder and decoder. Our approach, as a self-supervised task on target images, enables the encoder to acquire domain-invariant features and learn better target representations as shown in \cref{f1c}.
\item In Teacher-Student model, since the pseudo-label noise generated by the Teacher affects the Student model severely, we propose an adaptive confidence refinement module that progressively restricts the confidence metric for pseudo-label filtering. During the initial adaptation phase, soft confidence is applied allowing more pseudo-labels to learn better target representations. Subsequently, as confidence gradually increases, the focus shifts towards enhancing detection accuracy by prioritizing more reliable pseudo-labels.
\item We release a bounding box annotated subset of 1000 mammograms from the RSNA Breast Screening Dataset (referred to as \rsna) to support further research in \bcdm.
\item We setup a new state-of-the-art (\sota) in detection accuracy for \uda settings. We report a sensitivity of 0.74 on \inbreast~\cite{moreira2012inbreast} and 0.51 on \ddsm~\cite{lee2017curated} at 0.3 \fpi, compared to 0.61 and 0.44 using current \sota respectively. Significant performance gains are also observed on our in-house and \rsna datasets.
\end{enumerate*}

%
%
%
%

%% file: tables/fig1.tex
\begin{figure}[t]
    \centering
    \begin{subfigure}[b]{0.18\textwidth} 
        \centering
        \includegraphics[width=\textwidth, height=3cm]{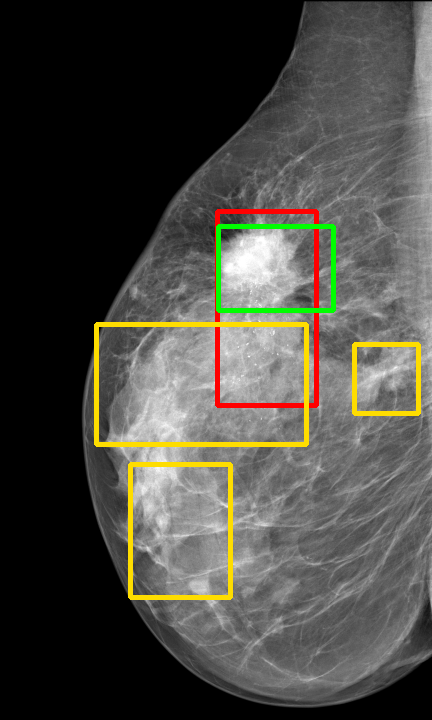}
        \subcaption{}
        \label{f1a}
    \end{subfigure} 
    \begin{subfigure}[b]{0.18\textwidth} 
        \centering
        \includegraphics[width=\textwidth, height=3cm]{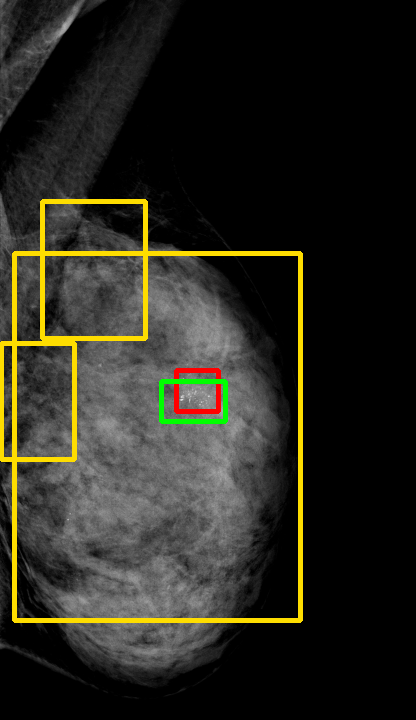}
        \subcaption{}
        \label{f1b}
    \end{subfigure}\hfill 
    \begin{subfigure}[b]{0.6\textwidth} 
        \centering
        \hspace{-0.8 cm}
        \includegraphics[width=\textwidth, height=3cm]{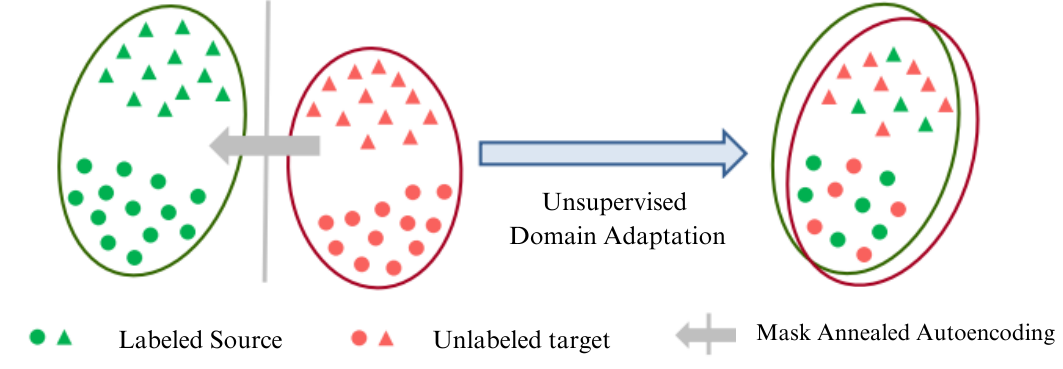}
        \subcaption{}
        \label{f1c}
    \end{subfigure}
    \caption{(a) and (b) depict false positive predictions by current teacher student models in cross-domain \bcdm. Red boxes indicate ground truth, yellow boxes show MRT \cite{zhao2023masked} predictions, and green boxes indicate predictions from \myname. As shown in (c), our approach effectively mitigates the domain gap and makes accurate predictions.}
    \label{fig:all}
\end{figure}

%% file: Sections/methodology.tex
\section{Proposed Method}
\mypara{Problem Formulation}
For a cross domain \bcdm problem in a \uda setting, we have an annotated source dataset $\D_s = \{(x_i^s, y_i^s)\}_{i=1}^{N_s}$ with $N_s$ samples, where each sample $x$ represents a mammogram, and $y = (b, c)$ denotes annotation for
malignancy, including bounding box $b$ and the corresponding malignant class
$c$. If the sample is benign, there is no corresponding bounding box annotation. Moreover, we have an unannotated target dataset $\D_t =  \{x_i^t\}_{i=1}^{N_t}$ with $N_t$ samples. The aim is to improve the performance on target data by training a model solely on source dataset $\D_s$ and target images $\D_t$ (target labels are not avaliable).
%
\begin{figure}[t]
    \centering
    \includegraphics[width=\textwidth]{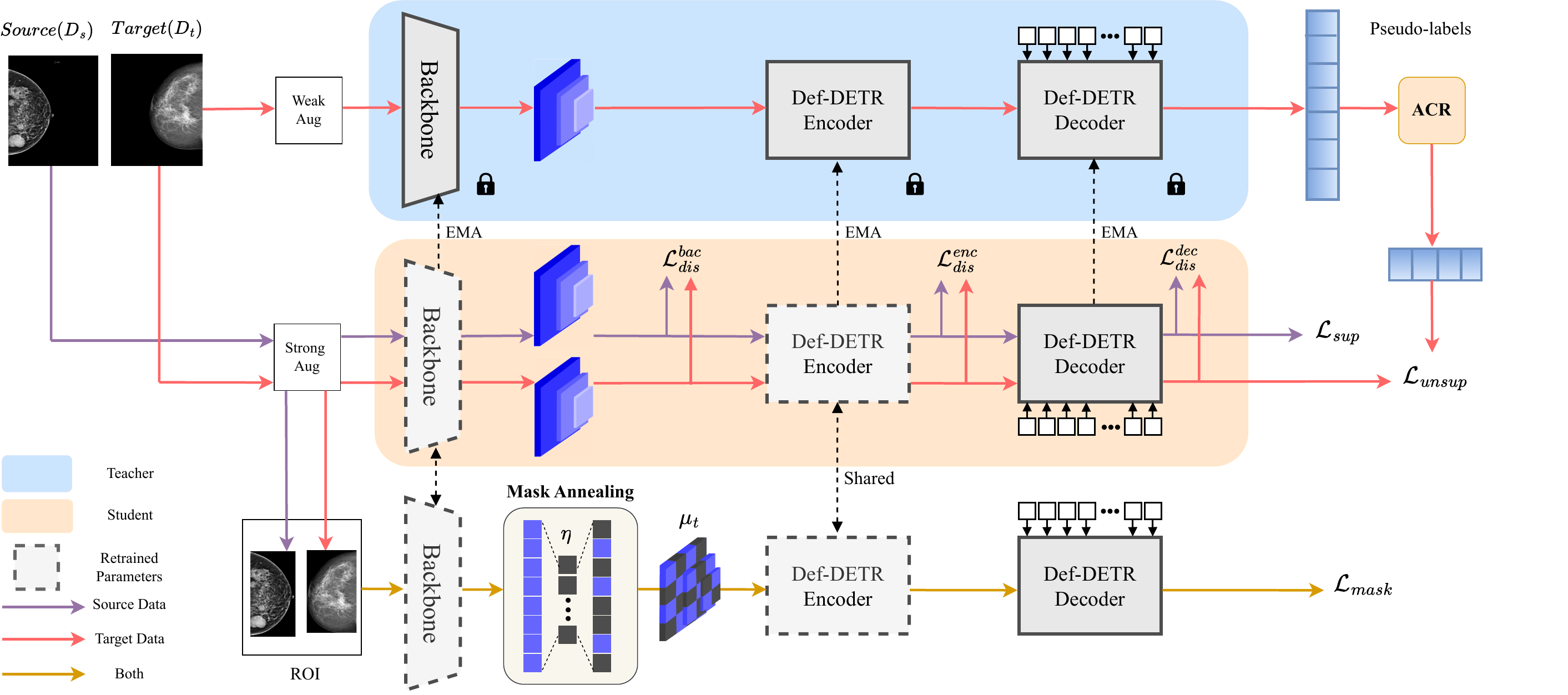}
   \caption{\small{Architecture of proposed Domain-invariant Mask Annealed Student Teacher Autoencoder (\myname) framework}}
    \label{f2}
\end{figure}
\subsection{Proposed Architecture}
Recent advances~\cite{li2022cross,yu2022cross,zhao2023masked} in \uda use an Adaptive Teacher-Student model with adversarial alignment and selective retraining to achieve domain adaptation. Our proposed \myname architecture follows the same style, and is shown in \cref{f2}. We utilize \emph{Deformable DETR} \cite{zhu2020deformable} (\defdetr) detector, pre-trained on the source domain, $\D_s$, as the backbone in our architecture. 
\mypara{Teacher-Student model}
The model comprises of two main branches: a target-specific Teacher and a cross-domain Student. The Teacher processes weakly augmented images exclusively from target domain ($D_t$), while the Student handles strongly augmented images from both domains ($D_s$ and $D_t$). Throughout the training process, the Teacher model generates pseudo-labels for $D_t$, which are then utilized to train the Student model. Additionally, the Student updates its acquired knowledge to the Teacher through exponential moving average (\ema) after each iteration, $\theta_T \leftarrow \alpha \theta_T + (1 - \alpha) \theta_S$. For source domain, supervised loss $\L_\text{sup}$ is calculated on $x_s$ using ground truth annotations same as~\cite{zhu2020deformable}, whereas for target domain $x_t$, we use unsupervised loss, $\L_\text{unsup}$, which is the cross-entropy loss with Teacher-generated pseudo-labels.
\mypara{Discriminators for adversarial alignment} 
As annotations are solely present for the source images $(\D_s)$, both the Teacher and the Student can be biased towards $\D_s$ during the learning process. To mitigate this~\cite{li2022cross,zhao2023masked} introduce adversarial learning in the Teacher-Student framework. To achieve adversarial learning, Domain discriminators $(D)$ are used after certain components to predict the domain label of the features, updated by binary cross-entropy loss, $\L_\text{dis}$. In \myname, the discriminators are placed after the backbone, \defdetr encoder, and decoder, as shown in \cref{f2}. We use standard adversarial loss, $ \L_\text{adv} = \mathop{\text{max}}_{\D_s} \mathop{\text{min}}_{\D_t} \L_{\text{dis}}$. Gradient Reverse Layers are used for min-max optimization. The overall objective of the Student $(\L_\text{teach})$ is the sum of supervised $(\L_\text{sup})$, unsupervised $(\L_\text{unsup})$, and adversarial $(\L_\text{adv})$ losses.
\mypara{Selective retraining}
Transformer-based models tend to over-fit on target data during such cross-domain training, especially when noisy annotations are included. As the Teacher model undergoes continuous updates via \ema from the Student, it could also be affected and produce incorrect and limited pseudo-labels. To address this, we follow~\cite{zhao2023masked}, and adopt selective retraining mechanism to help Student jump out of local optimums biased to wrong pseudo-labels. In \myname, the backbone and encoder of the Student are re-initialized with source-trained weights $\theta_{\text{s}}$ after certain epochs.

\subsection{Mask Annealed Autoencoder}

\input{algorithm}

\defdetr encoder in our proposed \myname architecture uses multi-scale feature maps $\{ X_i \in \mathbb{R}^{ C_i \times H_i \times W_i } \}_{i=1}^K$, where $K$ represents the number of feature map layers. We propose a novel mask annealing technique to mask the feature maps $\{ m_i \in \{0, 1\}^{H_i \times W_i} \}_{i=1}^K$ with initial masking ratio $\mu_t$, implying that $\mu_t \scriptsize{\%}$ of the pixels in the feature map are masked (set to zero). The easy-to-hard mask annealing curriculum, as shown in \cref{algo}\textbf{(left)}, is devised to adaptively mask patches to make the reconstruction task easier or difficult. During training, the step $\eta_t$ is optimized using stochastic gradient descent with warm restarts~\cite{loshchilov2016sgdr} as: 
\begin{equation}
    \eta_t = \eta^i_{\text{min}} + \frac{1}{2}(\eta^i_{\text{max}} - \eta^i_{\text{min}}) \left(1 + \cos\left(\frac{T_{\text{c}}}{T_{\text{i}}}\pi\right)\right). 
    \label{eq1}
\end{equation}
Here, $\eta^i_{\text{min}}$ and $\eta^i_{\text{max}}$ denote the ranges for the steps, $T_{\text{c}}$ reflects the number of iterations has been completed since the last restart and  $T_i$ is the no of iterations when a new warm restart of \texttt{SGD} is to be performed, 
\mypara{Reconstruction} 
Following the masking of feature maps, deformable attention is applied to further encode them by the \defdetr encoder~\cite{zhu2020deformable}. In the encoder's output features, we use a shared mask query $q_m$ to fill the masked portion and send it to the \mae decoder $D_s$ to reconstruct the masked portion. 
Given that the last layer of the feature maps, denoted as $X_K$, encapsulates all the semantic information~\cite{zhu2020deformable}, we solely reconstruct this layer to expedite faster convergence and lower computational overhead. The decoder's last layer consists of a linear projection, with output channels matching those of $X_K$. Finally, to supervise the reconstruction loss $\L_\text{mask}$, we compute the mean square error between output reconstructed feature maps $(\hat{X}_K)$ and original feature maps $(X_K)$. Hence, the overall objective of the Student model $\L$ is sum of the Teacher based loss $(\L_\text{teach})$, and the reconstruction loss $(\L_\text{mask})$. 
\input{tables/table0}
\mypara{Adaptive confidence refinement (ACR)} 
The correctness of the pseudo-labels is tightly related to confidence used for filtering. In the early stages of learning, confidence tends to be less reliable due to large domain shift. Therefore, we propose a gradual transition of confidence from soft to hard manner (\cref{algo} \textbf{(right)}). To this end, we start with the soft confidence $C_s$, and as iterations continue, we progressively assign more importance to hard confidence $C_h$ with a shifting weight $\delta$, s.t.: $C = (1 - \delta) \cdot C_s + \delta \cdot C_h$. The weight $\delta$ is determined as:
$
\delta = 2 \cdot \frac{1}{1 + \exp(-\alpha \frac{t}{e})} - 1, 
$
where, $t$ denotes current iteration, $e$ denotes total iterations,and $\alpha$ is a hyperparameter. At each iteration, a pseudo-label from the Teacher is considered valid (used to compute $\L_\text{unsup}$) if its confidence exceeds $C$.

%% file: algorithm.tex
\begin{figure*}[t]
\centering
\hspace{-5em}
\begin{minipage}[b][7.5cm][t]{0.60\linewidth} 
\begin{algorithm}[H]
\caption{Mask Annealing updating process}
\label{alg:adversarial_training}
\tiny
\begin{algorithmic}[1]
    \State \textbf{Input:} weight parameters ($\theta_T$), masking ratio ($\mu_t$), step ($\eta$), warm restart epoch ($T_{\text{i}})$, max epochs ($T_{\text{max}})$, epochs since last restart ($T_{\text{c}})$, ranges for the steps ($\eta^i_{\text{min}}$, $\eta^i_{\text{max}}$)
    \State \textbf{Data:} $b_{\text{st}} = \{(x^s_{i})\}_{i=1}^{N_s} \cup \{x^t_{j}\}_{j=1}^{N_t}$\hfill \textit{\Comment{Batch}}
    \State Initialize hyperparameters: $\mu = 0.3$, step $\eta = 0.01$, batch size $b = 16$, max epochs $T_{\text{max}} = 40$, initializing parameters of network $\theta_T$ using source trained parameters.
    
    \For{$i$ in $1$ to $T_{\text{max}}$}
        \State $\eta_t = \eta^i_{\text{min}} + \frac{1}{2}(\eta - \eta^i_{\text{min}}) \left(1 + \cos\left(\frac{T_{\text{c}}}{T_{\text{i}}}\pi\right)\right)$\hfill\textit{cosine annealing}
        
        \State  $L = \mathcal{L}_{\text{mask}}(b_{\text{st}}, \theta_T)$ \hfill \textit{\Comment{Computing loss for batch $b_{st}$}}

        \State  $\nabla_{\theta} L = \nabla_{\theta}f_t(b_{\text{st}}, \theta_T)$ \hfill \textit{\Comment{Computing gradients}}
        
        \State  $\theta_T' = \theta_T - \eta_t \cdot \nabla_{\theta} L$ \hfill \textit{\Comment{Updating parameters}}

       \State  $\mu = \mu + \eta_t$ if $L < \bar{L}$, otherwise $\mu = \mu - \eta_t$ \hfill \textit{\Comment{Updating $\mu$}}
        
        \State  $i = i + 1$ \hfill \textit{\Comment{Incrementing epoch}}
    \EndFor
    \State \textbf{Return} parameters $\theta_T'$, $\mu'$.
\end{algorithmic}
\end{algorithm}
\end{minipage}
\hspace{0.15em}
\begin{minipage}[b][6.7 cm][t]{0.22\linewidth} 
  \centering
  \includegraphics[height=5.0cm]{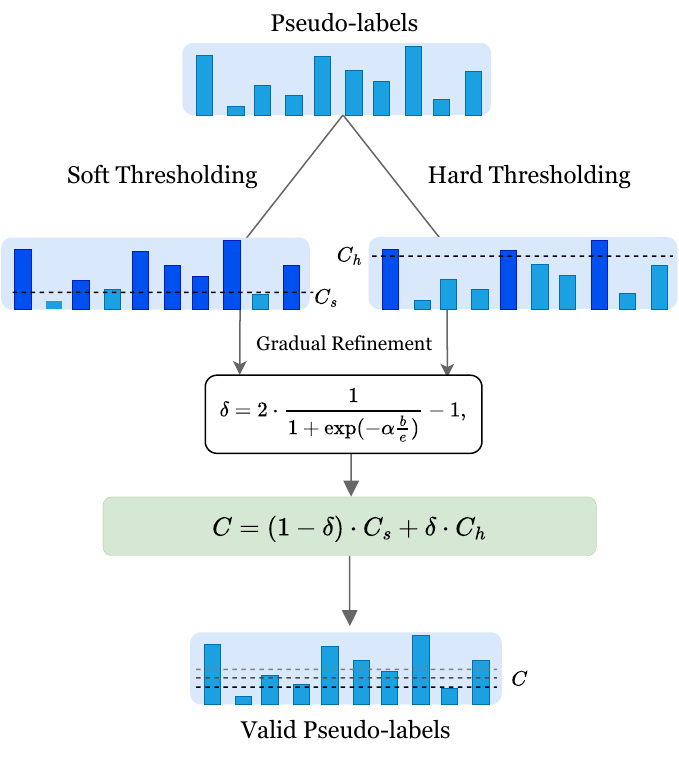} \\
  \label{fig:figure1and2}
\end{minipage}
\vspace{-3.5em}
\caption{Mask Annealing Algorithm \textbf{(left)} and Adaptive Confidence Refinement \textbf{(right)} flowchart depicts the gradual transition of confidence threshold from soft to hard.}
\label{algo}
\end{figure*}

%% file: tables/table0.tex
\begin{table*}[t]
\centering
\begin{minipage}[b]{0.68\linewidth} 
  \resizebox{\linewidth}{!}{
    \begin{tabular}{@{}c|l|l|cccccc|cc@{}}
    \toprule
    \textbf{Datasets}& \textbf{Model Name} & \textbf{Venue} & \textbf{R@0.05} & \textbf{R@0.1} & \textbf{R@0.3} &\textbf{R@0.5}& \textbf{R@1.0} & \textbf{R@2.0} & \textbf{Accuracy} & \textbf{F1-score} \\ \midrule
    & SFA\cite{wang2021exploring} & MM'21 &0.05 &0.12 & 0.19& 0.24 &0.31 &0.39 &0.336&0.287\\
    & UMT\cite{deng2021unbiased} & CVPR'21 &0.02 & 0.07& 0.11&0.16 &0.25 &0.37&0.308& 0.235\\
    & D-Adapt\cite{jiang2021decoupled} & ICLR'22 &0.08 & 0.13& 0.19&0.27 & 0.36&0.44&0.335 &0.215\\
    & AT\cite{li2022cross} & CVPR'22 &0.13 &0.18 & 0.22& 0.40& 0.52& 0.66&0.654&0.314 \\
    \textbf{\ddsm~\cite{lee2017curated} to} & H2FA\cite{xu2022h2fa} & CVPR'22&0.05 &0.09 & 0.15& 0.21& 0.28& 0.36& 0.291&0.225\\
    \textbf{\texttt{In-house}} & AQT\cite{huang2022aqt} & IJCAI'22 & 0.08& 0.11&0.18 & 0.26& 0.33&0.47&0.378& 0.306\\
    & HT\cite{deng2023harmonious}  & CVPR'23 &0.12 &0.17 & 0.32& 0.42&0.63 &0.75 &0.316&0.260 \\
    & ConfMIX\cite{mattolin2023confmix} & WACV'23 &0.09 &0.13 & 0.19& 0.30&0.46 &0.53 &0.517 &0.218\\
    & CLIPGAP\cite{vidit2023clip} & CVPR'23 &0.11 &0.14 & 0.31&0.40 &0.55 & 0.67&0.723&0.348 \\
    & MRT\cite{zhao2023masked} & ICCV'23 &0.13& 0.17&0.34&0.46 & 0.55& 0.70&0.816 &0.363  \\
    & \textbf{Ours} & & \textbf{0.19} & \textbf{0.26}&\textbf{0.43} & \textbf{0.54} & \textbf{0.68} & \textbf{0.78} & \textbf{0.835} & \textbf{0.392} \\ \midrule
    & SFA~\cite{wang2021exploring} & MM'21& 0.03&0.04 &0.13 & 0.23& 0.35&0.53&0.241&0.338\\
    & UMT\cite{deng2021unbiased} & CVPR'21 & 0.01&0.05 &0.09 &0.15 & 0.21&0.25 &0.204&0.319\\
    & D-Adapt\cite{jiang2021decoupled} & ICLR'22&0.11 & 0.23&0.29 &0.45 &0.59 &0.65 &0.731&0.414\\
    & AT\cite{li2022cross} & CVPR'22 &0.19 &0.27 &0.38 &0.65 &0.75 &0.78 &0.721&0.512\\
    \textbf{\texttt{\ddsm~\cite{lee2017curated}} to} & H2FA\cite{xu2022h2fa} & CVPR'22& 0.13&0.20 &0.32 &0.45 &0.52 &0.58 &0.575&0.312\\
    \textbf{\inbreast~\cite{moreira2012inbreast}} & AQT\cite{huang2022aqt} & IJCAI'22 &0.11 &0.24 & 0.37&0.44 &0.57 &0.64 &0.680&0.349\\
    & HT\cite{deng2023harmonious}  & CVPR'23 &0.17 &0.31 &0.49 &0.61 &0.69 &0.73 &0.704& 0.362\\
    & ConfMIX\cite{mattolin2023confmix} & WACV'23& 0.19& 0.29&0.47 &0.58 &0.73 &0.75 &0.737&0.409 \\
    & CLIPGAP\cite{vidit2023clip} & CVPR'23 &0.15 &0.30 &0.55 & 0.61&0.75 &0.79 &0.712&0.445\\
    & MRT\cite{zhao2023masked} & ICCV'23 &0.16 &0.31 &0.54 & 0.64&0.72 & 0.77&0.789&0.489\\
    & \textbf{Ours} & & \textbf{0.25} & \textbf{0.3} & \textbf{0.61}&\textbf{0.70} & \textbf{0.82} & \textbf{0.82} & \textbf{0.808} & \textbf{0.524} \\ \midrule
    & SFA~\cite{wang2021exploring} & MM'21& 0.01&0.03 &0.09 &0.16 &0.19 &0.22&0.425&0.271\\
    & UMT\cite{deng2021unbiased} & CVPR'21 & 0.05& 0.09& 0.11& 0.15& 0.21&0.26 &0.362&0.216\\
    & D-Adapt\cite{jiang2021decoupled} & ICLR'22& 0.04& 0.06& 0.12& 0.18&0.29 &0.36 &0.471&0.263\\
    & AT\cite{li2022cross} & CVPR'22 &0.16 & 0.21&0.28 &0.35 &0.42 &0.48 &0.712&0.338\\
    \textbf{\texttt{In-house} to} & H2FA\cite{xu2022h2fa} & CVPR'22& 0.03& 0.07&0.13 & 0.18&0.26 & 0.35&0.486&0.236\\
    \textbf{\inbreast~\cite{moreira2012inbreast}} & AQT\cite{huang2022aqt} & IJCAI'22 &0.02 &0.10 &0.17 &0.25 &0.33 &0.39 &0.569&.318\\
    & HT\cite{deng2023harmonious}  & CVPR'23 &0.07 &0.11 & 0.18&0.26 & 0.31&0.38 &0.648&0.297 \\
    & ConfMIX\cite{mattolin2023confmix} & WACV'23& 0.03& 0.08&0.15 & 0.23& 0.28&0.32 &0.527&0.285 \\
    & CLIPGAP\cite{vidit2023clip} & CVPR'23 &0.06 &0.13 &0.19 &0.28 &0.36 & 0.41&0.621&0.310 \\
    & MRT\cite{zhao2023masked} & ICCV'23 &0.32 &0.43&0.52 &0.69 & 0.72&0.78 &0.779&0.352\\
    & \textbf{Ours} & & \textbf{0.46} & \textbf{0.53}&\textbf{0.71} & \textbf{0.75} & \textbf{0.79} & \textbf{0.82} & \textbf{0.812}& \textbf{0.395} \\ \midrule
    & SFA\cite{wang2021exploring} & MM'21& 0.02&0.05 & 0.09& 0.12& 0.15& 0.19&0.501&0.430\cr
    & UMT\cite{deng2021unbiased} & CVPR'21  & 0.04& 0.06& 0.10& 0.13& 0.19& 0.21&0.528&0.287\cr
    & D-Adapt\cite{jiang2021decoupled} & ICLR'22& 0.03& 0.07& 0.09& 0.14& 0.18&0.27 &0.518&0.329 \cr
    & AT\cite{li2022cross} & CVPR'22  &0.18 & 0.24& 0.29&0.36 &0.39 &0.48 &0.698&0.601\cr
    \textbf{\texttt{In-house} to} & H2FA\cite{xu2022h2fa} &CVPR'22&0.02 &0.07 & 0.13& 0.19& 0.23& 0.28& 0.632& 0.325\cr
    \textbf{\rsna} & AQT\cite{huang2022aqt} &IJCAI'22& 0.07& 0.12& 0.17& 0.25& 0.31&0.38 &0.574 & 0.415\cr
    & HT\cite{deng2023harmonious}  & CVPR'23 &0.09& 0.15& 0.19& 0.28&0.37 &0.45 & 0.621&0.539\cr
    & ConfMIX\cite{mattolin2023confmix} & WACV'23 &0.18&0.21 & 0.28&0.33 &0.41 &0.48 & 0.578& 0.413\cr
    & CLIPGAP\cite{vidit2023clip} & CVPR'23 &0.12&0.18 & 0.23&0.27 & 0.36& 0.42&0.769 &0.586\cr
    & MRT\cite{zhao2023masked} & ICCV'23 &0.28 & 0.37& 0.44& 0.53& 0.68&0.71 &0.812 & 0.524\cr
    & \textbf{Ours} & &  \textbf{0.36} & \textbf{0.41} & \textbf{0.58} & \textbf{0.65} & \textbf{0.70} & \textbf{0.79} & \textbf{0.888}&\textbf{0.653} \cr \bottomrule
    \end{tabular}
  }
  \vspace{0.2em}
  \caption{Comparison with state of the art \uda techniques.}
  \label{tab:sota_uda_comparison}
\end{minipage}
\hfill
\begin{minipage}[b][\totalheight][b]{0.3\linewidth} 
  \centering
  \includegraphics[width=\linewidth,height=0.63\linewidth]{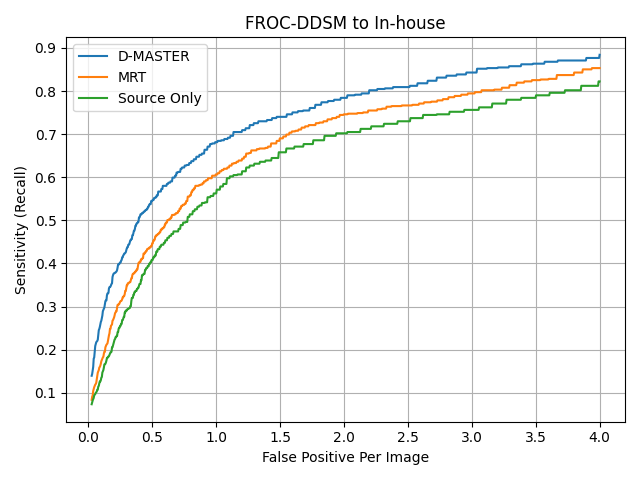} \\
  \includegraphics[width=\linewidth,height=0.63\linewidth]{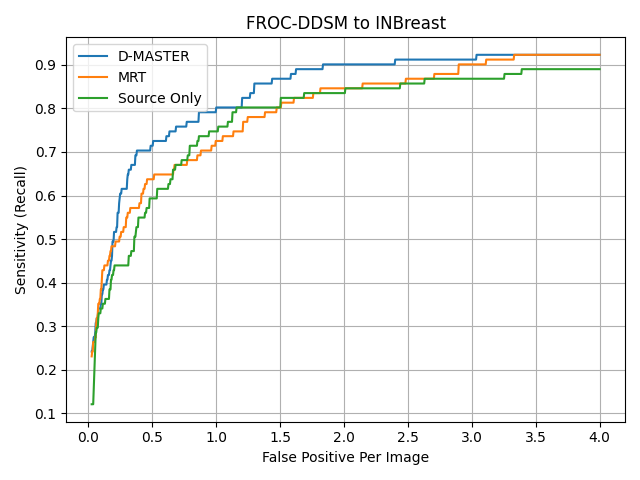}\\
  \includegraphics[width=\linewidth,height=0.63\linewidth]{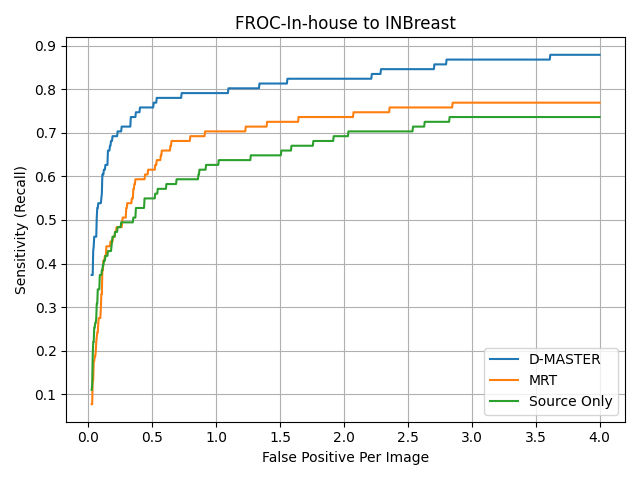} \\
  \includegraphics[width=\linewidth,height=0.63\linewidth]{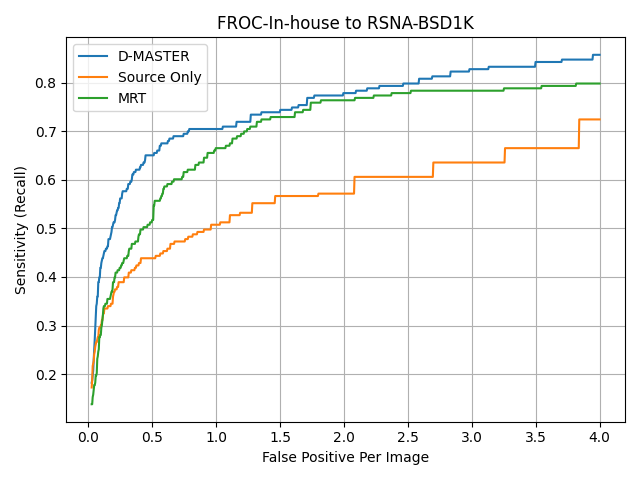} \\
  \vspace{0.3em}
  \captionof{figure}{\froc curves}
  \label{fig:sota_uda_froc}
\end{minipage}
\end{table*}

%% file: Sections/results.tex
\section{Experiments and Results}
\myfirstpara{Dataset and evaluation methodology}
We evaluate our proposed \myname on \inbreast \cite{moreira2012inbreast}, \ddsm \cite{lee2017curated}, \texttt{RSNA} \cite{carrrsna}, and in-house datasets. Our in-house diagnostic dataset contains 3,501 mammograms, including both CC and MLO views, with 615 malignancies. The original \texttt{RSNA} dataset \cite{carrrsna} consists of 54,706 screening mammograms, containing 1,000 malignancies from 8,000 patients. We curated a subset named \rsna, comprising 1,000 mammograms with 200 malignant cases, annotated at the bounding box level by 2 expert radiologists. Note that unlike single domain detection techniques which use a particular subset of the dataset for training and remaining for testing, our technique does not require any labels from the target dataset. Hence, for our problem, it seems logical to use the whole dataset during training and testing, and not just the respective train or test split. Hence, when reporting results for ``Dataset A to Dataset B'', we imply that the model is trained on $\D_s=A$ (whole dataset for the training), and adapted for $\D_t=B$ (whole dataset for \uda and testing). 
\input{tables/table2}

\mypara{Evaluation metric}
We use Free-Response Receiver Operating Characteristic (\froc) curves~\cite{egan1961operating} for reporting our results. The curves provide a graphical representation of sensitivity/recall values at different false positives per image (\fpi). We follow related works in this area~\cite{rangarajan2023deep} and consider a prediction as true positive if the center of the predicted bounding box lies within the ground-truth box. 

\mypara{Implementation details}
We employ a 2-layer asymmetric decoder~\cite{he2022masked} in the \mae with an initial mask ratio of 0.2. Network optimization uses the Adam optimizer~\cite{kingma2014adam} with an initial learning rate of $2 \times 10^{-4}$, and a batch size of 16. Data augmentation techniques include random horizontal flips, high-resolution mammograms (4K) for weak augmentation, Gaussian blurring, low-resolution mammograms (1K), and random contrast change for strong augmentations. 
Detailed information is in supplementary \cref{tab:implementation_details}.

\mypara{Comparison with \sota techniques}
\cref{tab:sota_uda_comparison} shows the comparative results with other domain adaptation techniques, including those proposed for natural images. \cref{fig:sota_uda_froc} depicts corresponding \froc curves comparison with the nearest competitors only (to avoid clutter). \cref{tab:sota_uda_comparison_suppl} in the supplementary show quantitative results for more datasets.


\input{tables/table-ablation}
\mypara{Ablation Study}
Our ablation study include three experiments:
\begin{enumerate*}[label=\textbf{(\arabic*)}]
\item \cref{tab:ablation} provides the quantitative and qualitative analysis to understand the impact of various proposed components in the proposed \myname architecture. 
\item Our core contributions are not specific to medical imaging, and are expected to be useful for natural images as well. Hence, we compare our model with existing domain adaptive object detection (DOAD) techniques when trained on \texttt{Sim10K}~\cite{johnson2016driving} data and unsupervised domain adaptation on \texttt{Cityscapes}~\cite{cordts2016cityscapes} dataset. Since our model is designed for a single label (presence or absence of breast cancer), hence we perform this experiment only for single label, ``car'' on the two datasets. \cref{ablation}\textbf{(left)} shows the result.
\item To understand the benefit of overall solution with respect to recent \sota object detection techniques, in \cref{ablation}\textbf{(right)}, we show the result when these techniques are trained on in-house dataset and directly tested on \inbreast without any domain adaptation.
\end{enumerate*}

%% file: tables/table2.tex
\begin{table}[t]
\centering
\resizebox{\textwidth}{!}{%
	\setlength{\tabcolsep}{6pt}
\begin{tabular}{c|c|c|c|cccccc}
\hline
\textbf{Source} & \textbf{AT~\cite{li2022cross}} & \textbf{MA} & \textbf{ACR} & \textbf{R@0.025} & \textbf{R@0.05} & \textbf{R@0.1} & \textbf{R@0.3} & \textbf{R@0.5} & \textbf{R@0.9} \\
\hline
\checkmark & & & & 0.19 & 0.34 & 0.43 & 0.57 & 0.65 & 0.69 \\
\checkmark & & \checkmark & & \textbf{0.39} & 0.46 & 0.53 & 0.71 & 0.75 & 0.78 \\
\checkmark & \checkmark & & & 0.28 & 0.35 & 0.42 & 0.61 & 0.64 & 0.69 \\
\checkmark & \checkmark & & \checkmark & 0.30 & 0.42 & 0.51 & 0.61 & 0.70 & 0.71 \\
\checkmark & \checkmark & \checkmark & & 0.37 & 0.46 & 0.59 & 0.72 & 0.75 & 0.75 \\
\checkmark & \checkmark & \checkmark & \checkmark & \textbf{0.39} & \textbf{0.51} & \textbf{0.53} & \textbf{0.71} & \textbf{0.75} & \textbf{0.79} \\
\hline
\end{tabular}
}%
\\ \vspace{0.5em}\begin{tabular}{cccccc}
\includegraphics[width=0.158\textwidth, height= 2.7 cm]{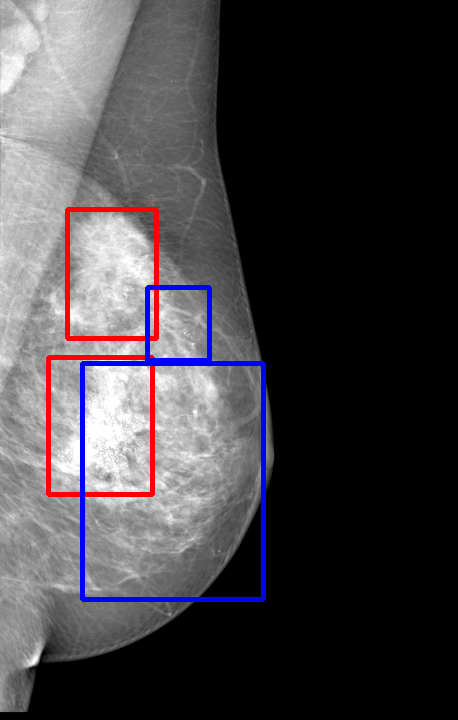}
&
\includegraphics[width=0.158\textwidth, height= 2.7 cm]{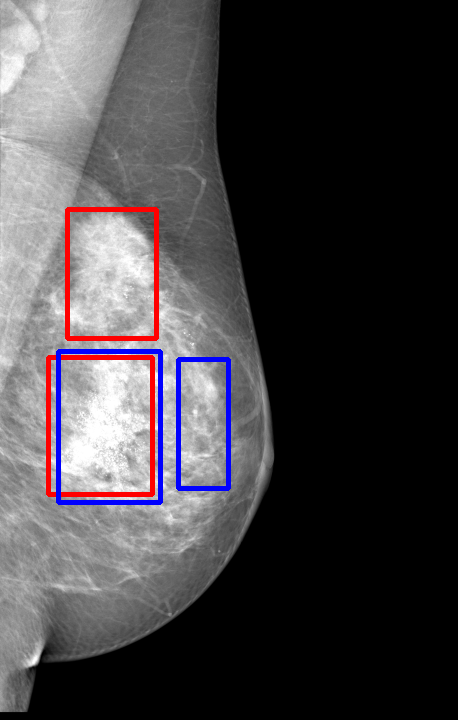}
&
\includegraphics[width=0.158\textwidth, height= 2.7 cm]{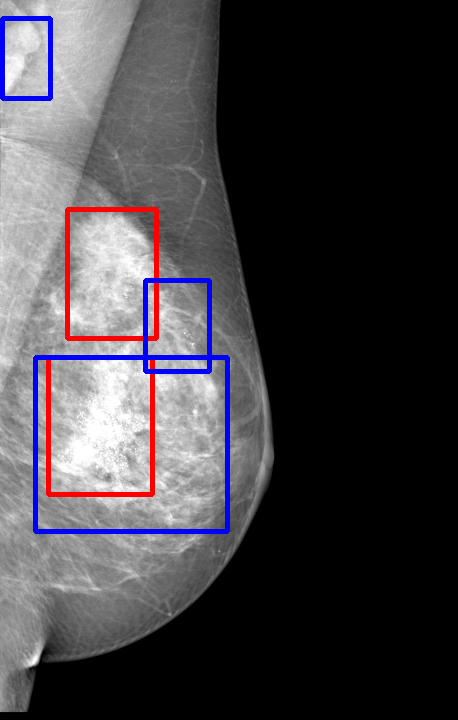}
&
\includegraphics[width=0.158\textwidth, height= 2.7 cm]{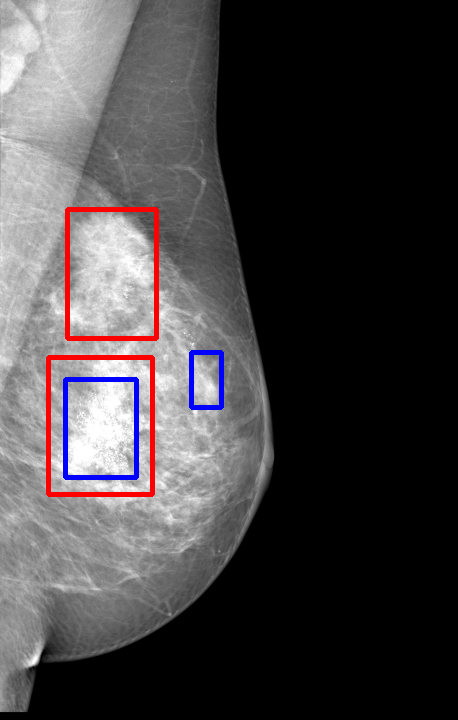}
&
\includegraphics[width=0.158\textwidth, height= 2.7 cm]{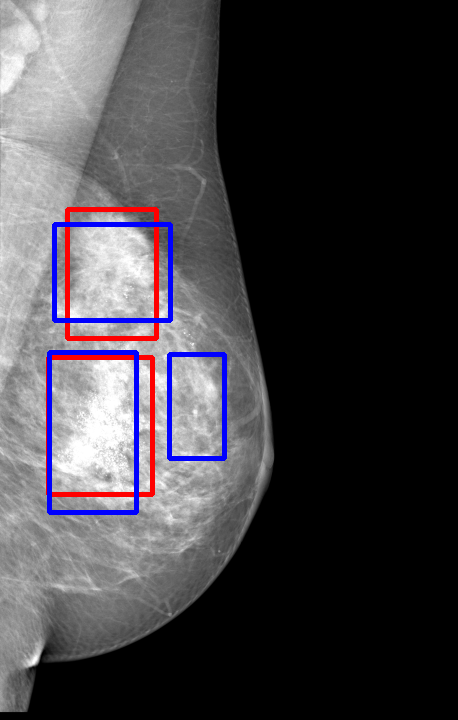}
&
\includegraphics[width=0.158\textwidth, height= 2.7 cm]{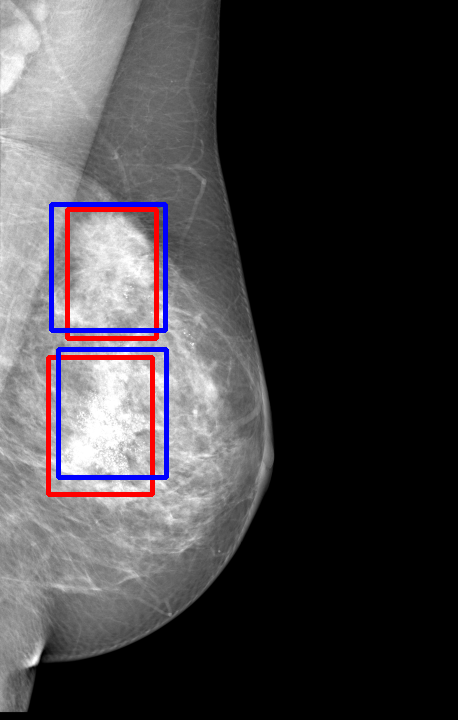}
\end{tabular}
\caption{\textbf{Ablation study to understand impact of each proposed module for In-house to \inbreast adaptation.} ``Source'' denotes the source-only trained model, ``Baseline'' the basic teacher-student architecture, ``MA'' the proposed mask annealing technique, and ``ACR'' denotes adaptive confidence refinement module. The figures from left to right correspond to qualitative results from row 1 to row 6 respectively. Red boxes denote the ground truth, and blue boxes show the predicted regions.}
\label{tab:ablation}
\end{table}

%% file: tables/table-ablation.tex
\begin{table*}[t]
\caption{
	\textbf{Left:} Demonstrating superiority of our approach for natural image setting with \texttt{Sim10K}~\cite{johnson2016driving} as the source, and \texttt{Cityscapes}~\cite{cordts2016cityscapes} as the target dataset.
	\textbf{Right:} Comparing object detectors proposed for natural images with our method.  
}
\vspace{0.2em}
\centering
\label{ablation}
\begin{adjustbox}{width=\textwidth}
\begin{tabular}{@{}l|l|ccc|c|ccl|l|c@{}}

         \cline{1-7} \cline{9-11}
        \makecell[c]{\textbf{Detector}} & \makecell[c]{\textbf{Venue}}  & \textbf{R@0.1} & \textbf{R@0.3} & \textbf{R@0.5}  & \hspace{0.1 cm}\textbf{Acc} \hspace{0.1 cm} & \hspace{0.1 cm}\textbf{F1} & \hspace{0.5cm}  & \makecell[c]{\textbf{Detector}} & \makecell[c]{\textbf{Venue}} & \textbf{AP(car)}
        \\
        \cline{1-7} \cline{9-11}
        FRCNN~\cite{ren2015faster} & NIPS'15   & 0.25 & 0.40 & 0.45 & 0.828 &0.297&  &FRCNN~\cite{ren2015faster}     & NIPS'15            & 39.4    \\
        YOLO-v5~\cite{ge2021yolox}& ICCVW’21    & 0.26 & 0.34 & 0.36 & 0.828 & 0.311& &SSAL~\cite{munir2021ssal} & NIPS'21        & 42.5         \\
        Cond.DETR~\cite{ge2021yolox} & ICCV’21  & 0.30 & 0.47 & 0.51  &0.833  &0.430 & &DA-Faster~\cite{chen2018domain}        & CVPR'18            & 41.9 \\   
        Dab-DETR~\cite{liu2022dab} & ICLR’22   & 0.43 & 0.57 & 0.63  & 0.843 &0.552 & & MeGACDA~\cite{vs2021mega}        & CVPR'21            & 44.8 \\   
        Dab DEF~\cite{liu2022dab}  & ICLR’22   & 0.40 & 0.46 & 0.50  & 0.830 & 0.546&& ViSGA~\cite{rezaeianaran2021seeking}            & ICCV'21            & 49.3 \\
        DN-DETR~\cite{li2022dn} & CVPR’22  & 0.38 & 0.47 & 0.58  & 0.828 & 0.393& &AQT~\cite{huang2022aqt} & IJCAI 2022          & 53.4          \\
        DN DEF.~\cite{li2022dn} & CVPR’22  & 0.39 & 0.44 & 0.47  & 0.825 & 0.526& &DA-Detr~\cite{zhang2023detr}   & CVPR'23          & 54.7        \\
        YOLO-V8~\cite{jocher2023yolo} & UT’23  & 0.06 & 0.08 & 0.10  & 0.258 &0.346 & & KTNet~\cite{tian2021knowledge}            & ICCV'21            & 50.7  \\
        FocalNET~\cite{yang2022focal} & NIPS’23   & 0.56 & 0.69 & 0.74  & 0.865 &0.641& &ConfMix~\cite{mattolin2023confmix} & CVPR'23          & 56.3       \\
        DINO~\cite{zhang2023dino} & ICLR’23   & 0.35 & 0.54 & 0.65 & 0.820 & 0.275& &PT~\cite{chen2022learning}               & PMLR'22            & 55.1  \\            
        \textbf{Ours} &  - & \textbf{0.53} & \textbf{0.71}  & \textbf{0.75}  & \textbf{0.879} &  \textbf{0.773}& & \textbf{Ours}  & -          & \textbf{61.9}        \\
        \cline{1-7} \cline{9-11}
    \end{tabular}%
\end{adjustbox}

\end{table*}

%% file: Sections/conclusion.tex
\section{Conclusion}
Lack of generalisability of Deep Neural Networks for breast cancer detection, on the data obtained across different geographies, with different distributions, acquired on different machines and imaging protocols, is a big barrier towards their clinical adoption. In this paper, we address this problem by proposing a new unsupervised domain adaptation framework, which can help such models adapt to newer sample distributions. Experimental results, carried out with four datasets from different domains, demonstrated the superior performance over other competing methods. We hope that our work will improve the clinical adoption of automated breast cancer detection models with improved generalization using our technique.

%% file: Sections/acknowledgment.tex
\begin{credits}
\subsubsection{\ackname} We acknowledge and thank the funding support from Ministry of Education, Government of India, Central Project Management Unit, IIT Jammu with sanction number IITJMU/CPMU-AI/2024/0002.

\subsubsection{Disclosure of Interests.} The authors have no competing interests to declare relevant to this article's content.
\end{credits}

%% file: Sections/supplementary.tex
\begin{center}
{\Large Supplementary material for Paper ID: 1343}
\end{center}

\input{tables/table3}
\input{tables/table6} 
\input{tables/table7} 

\input{tables/table8} 

%% file: tables/table3.tex
    \begin{table}[H]
        \begin{minipage}{0.65\textwidth}
            \begin{minipage}{0.50\textwidth}
                \centering
                \begin{subtable}{\textwidth}
                    \centering
                    \caption{Mask Ratio}
                    \begin{tabular}{cc}
                        \toprule
                        Mask Ratio & R@0.3 \\
                        \midrule
                        0.3   & \textbf{0.61}  \\
                        0.5   & 0.57  \\
                        0.8  & 0.44   \\
                        \bottomrule
                    \end{tabular}
                \end{subtable}
                
                \begin{subtable}{\textwidth}
                    \centering
                    \caption{Coefficient Decay}
                    \begin{tabular}{cc}
                        \toprule
                        Strategy & R@0.3 \\
                        \midrule
                        No decay   & 0.65   \\
                        Linear   & 0.72   \\
                        Hard   & \textbf{0.74}   \\
                        \bottomrule
                    \end{tabular}
                \end{subtable}

            \end{minipage}%
            \begin{minipage}{0.50\textwidth}
                \centering
                \begin{subtable}{\textwidth}
                    \centering
                    \caption{Training Data}
                    \begin{tabular}{cc}
                        \toprule
                        Training Data & R@0.3 \\
                        \midrule
                        Source   & 0.63 \\
                        Target   &  0.71 \\
                        Source and Target & \textbf{0.74}  \\
                        \bottomrule
                    \end{tabular}
                \end{subtable}
                 
                \begin{subtable}{\textwidth}
                    \centering
                    \caption{Annealing}
                    \begin{tabular}{cc}
                        \toprule
                        Annealing  & R@0.3 \\
                        \midrule
                        Step   & 0.   \\
                        Cyclic   & 0.71   \\
                        Cosine  & \textbf{0.74}   \\
                        \bottomrule
                    \end{tabular}
                \end{subtable}

            \end{minipage}
        \end{minipage}%
        \begin{minipage}{0.35\textwidth}
            \centering
            
            \includegraphics[width=\textwidth]{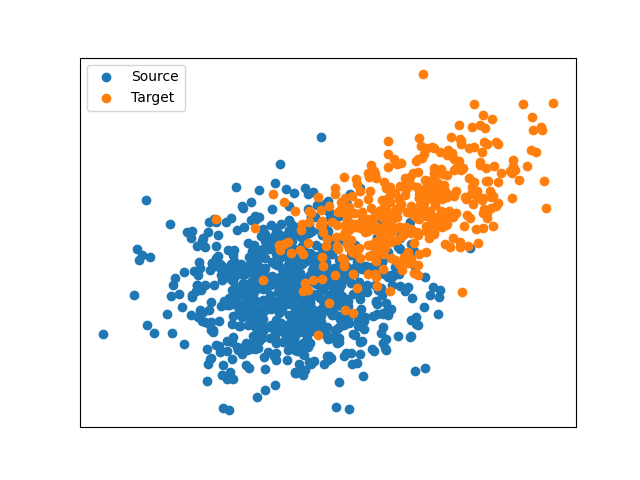} \\
            \includegraphics[width=\textwidth]{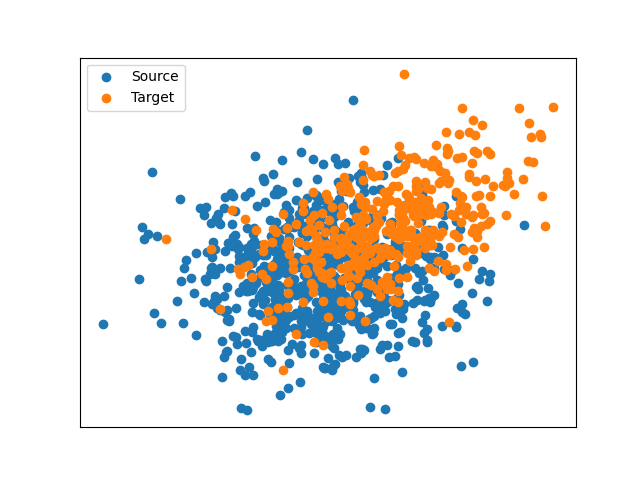}
            \label{fig:figure1and2}
        \end{minipage}
        \caption{ \small{The MAE ablation study shows that a low initial mask ratio, decaying $\lambda_{mask}$, and training on both source and target features improve performance, with D-MASTER reducing overfitting.}}
    \end{table}

%% file: tables/table6.tex
\begin{table}[h]
    \centering
   
    \label{tab:implementation_details}
    \begin{adjustbox}{width= 0.89\textwidth, height= 0.25\textwidth}
    \begin{tabular}{@{}clcc@{}}
        \toprule
        \textbf{Hyper-parameter} & \textbf{Description} & \textbf{INBreast} & \textbf{DDSM}  \\
        \midrule
       $N_c$ & Number of categories for classification head  & 2 & 2  \\
        $N_l^{\text{enc}}$& Number of encoder layers  & 8 & 6  \\
        $N_l^{\text{dec}}$ & Number of decoder layers   & 8 & 6   \\
        $N_l^{\text{aux}}$ &Number of MAE auxiliary decoder layers  & 2& 2  \\
        $N_q^{\text{dec}}$ & Number of queries for decoder   & 300 & 300 \\
        $N_q^{\text{aux}}$ &Number of queries for MAE auxiliary decoder & 882  & 882\\
        $H$& Number of hidden dimension for deformable attention& 256 &  256\\
        $F$ & Number of feedforward dimension for deformable attention   & 1024 &1024 \\
        $L$ & Number of feature levels for deformable attention   & 4& 4  \\
        $M$ & Number of heads for deformable attention & 8& 8\\
        $K$ & Number of reference points for each attention head& 4 & 4 \\
        $B$ & Batch Size during training   & 16 &16   \\
        $lr$ & Learning rate for modules except backbone and projection   & $2 \times 10^{-4}$&$ 2 \times 10^{-4}$\\
        $lr_{bac}$ & Learning rate for backbone and projection modules   & $2 \times 10^{-5}$ & $2 \times 10^{-5}$ \\
        $\beta_{bac}$ & Coefficient of discrimination loss after backbone $L_{\text{bac}}^{\text{dis}}$& 0.3 & 0.3 \\
        $\beta_{enc}$ & Coefficient of discrimination loss after encoder $L_{\text{enc}}^{\text{dis}}$
 & 1.0 &  1.0 \\
        $\beta_{dec}$ &Coefficient of discrimination loss after decoder $L_{\text{enc}}^{\text{dis}}$
 &  1.0 &  1.0 \\
        $\lambda_{unsup}$ & Coefficient of unsupervised loss $\mathcal{L}_{unsup}$ &  1.0 &  1.0 \\
        $\lambda_{mask}$ & Coefficient of supervised loss $\mathcal{L}_{sup}$ &  1.0 &  1.0\\
        $\gamma$ & EMA update ratio & 0.9996 & 0.9996 \\
        $\mu_{t}$ & Initial Mask ratio in MAE branch & 0.2 & 0.3 \\
        $\eta$ & Initial step for annealing & 0.2 & 0.3 \\

        $\eta^i_{min}$ & Minimum jump in mask annealing step $\eta$ & 0.05 & 0.05 \\
        $\eta^i_{max}$ & Maximum jump in mask annealing step $\eta$ & 0.15 & 0.15 \\
        $C_s$&Soft Confidence Metric&0.15&0.20\\
        $C_h$&Hard Confidence Metric&0.80&0.90\\\
        $E_{pre}$ & MAE branch with source data training epoch number & 88 & 87  \\
        $E_{teach}$ & Teacher-student training epoch number & 84 & 76 \\
        $E_{decay}$ & After Edecay epochs in teaching stage, we drop the MAE branch & 30 & 10\\
        $E_{reinit}$ & Re-initialization epoch for selective retraining & 40 & 20 \\
        
        \bottomrule
    \end{tabular}
    \end{adjustbox}
    \vspace{0.3 cm}
    \caption{Below are the detailed hyper-parameters corresponding to each benchmark, with the source dataset as the In-house dataset}
\end{table}

%% file: tables/table7.tex
\begin{table*}[t]
\centering
  \resizebox{\linewidth}{!}{
    \begin{tabular}{@{}c|l|l|cccccc|cc@{}}
    \toprule
    \textbf{Datasets}& \makecell[c]{\textbf{Model Name}} & \makecell[c]{\textbf{Venue}} & \textbf{R@0.05} & \textbf{R@0.1} & \textbf{R@0.3} &\textbf{R@0.5}& \textbf{R@1.0} & \textbf{R@2.0} & \textbf{Accuracy} & \textbf{F1-score} \\ \midrule
    & SFA\cite{wang2021exploring} & MM'21&0.01 & 0.01& 0.05& 0.07& 0.11&0.313 & 0.216&0.329\cr
    & UMT\cite{deng2021unbiased} & CVPR'21 &0.0 & 0.01&0.04 & 0.07& 0.09&0.13 & 0.261&0.362\cr
    & D-Adapt\cite{jiang2021decoupled} & ICLR'22 &0.0 & 0.02& 0.06&0.09 &0.10 &0.13 &0.382 &0.215\cr
    & AT\cite{li2022cross} & CVPR'22 & 0.01& 0.03&0.08 &0.10 &0.15 & 0.21&0.216 & 0.311\cr
    \textbf{\texttt{In-house} to} & H2FA\cite{xu2022h2fa} & CVPR'22 &0.02 & 0.03&0.06 &0.10 & 0.12&0.17 &0.371&0.315 \cr
    \textbf{\ddsm~\cite{lee2017curated}} & AQT\cite{huang2022aqt} & IJCAI'22 &0.01 &0.03 & 0.07& 0.13& 0.15&0.18 &0.412&0.398\cr
    & HT\cite{deng2023harmonious}  & CVPR'23 & 0.03&0.05&0.08 &0.10 &0.13 & 0.15&0.362&0.362 \cr
    & ConfMIX\cite{mattolin2023confmix} & WACV'23& 0.02&0.04 &0.09 &0.12 &0.16 &0.19 &0.336 &0.412\cr
    & CLIPGAP\cite{vidit2023clip} & CVPR'23 &0.01& 0.03& 0.07& 0.11& 0.15& 0.16& 0.336& 0.458\cr
    & MRT\cite{zhao2023masked} & ICCV'23 &0.03& 0.04& 0.09&0.12 & 0.17& 0.21&0.421 &0.587 \cr
    & \textbf{Ours} &- & \textbf{0.02}& \textbf{0.05} & \textbf{0.12} & \textbf{0.17} & \textbf{0.29} & \textbf{0.49}&\textbf{0.561}&\textbf{0.613} \cr \midrule
    & SFA\cite{wang2021exploring} & MM'21& 0.03& 0.07&0.13 &0.18 & 0.27&0.31& 0.629&0.210\cr
    & UMT\cite{deng2021unbiased} & CVPR'21  & 0.01&0.04& 0.09&0.15 &0.19 &0.23 &0.568&0.193\cr
    & D-Adapt\cite{jiang2021decoupled} & ICLR'22&0.03 &0.06 & 0.11& 0.18&0.25 &0.31 &0.668&0.241 \cr
    & AT\cite{li2022cross} & CVPR'22  & 0.10&0.28& 0.37& 0.45& 0.51& 0.66&0.725&\textbf{0.319}\cr
    \textbf{\rsna to} & H2FA\cite{xu2022h2fa} &CVPR'22& 0.03& 0.06& 0.14& 0.17&0.21 & 0.24& 0.591&0.274 \cr
    \textbf{\texttt{In-house}} & AQT\cite{huang2022aqt} &IJCAI'22&0.01 &0.05 &0.08 & 0.11& 0.17& 0.20& 0.527&0.230 \cr
    & HT\cite{deng2023harmonious}  & CVPR'23 &0.02& 0.10& 0.17&0.24 &0.33 & 0.41& 0.710&0.291\cr
    & ConfMIX\cite{mattolin2023confmix} & WACV'23 &0.03& 0.09&.0.16 &0.28 & 0.35& 0.39& 0.622&0.263 \cr
    & CLIPGAP\cite{vidit2023clip} & CVPR'23 &0.04& 0.08&0.23 & 0.36&0.64 &0.72 & 0.797&0.231\cr
    & MRT\cite{zhao2023masked} & ICCV'23 &0.06 & 0.11&0.29 &0.44 & 0.65&0.76 & 0.825& 0.304\cr
    & \textbf{Ours} & &  \textbf{0.14} & \textbf{0.20} & \textbf{0.37} & \textbf{0.54} & \textbf{0.72} & \textbf{0.83} & \textbf{0.825}&0.312 \cr \bottomrule
    \end{tabular}
  }
  \vspace{0.2em}
  \caption{\cref{tab:sota_uda_comparison} in the main paper showed similar comparison with \sota \uda methods on few dataset pairs. Here we show results for few more pairs.}
  \label{tab:sota_uda_comparison_suppl}

\end{table*}

%% file: tables/table8.tex
\begin{figure}[htbp]
\centering
\begin{subfigure}[b]{0.15\textwidth}
  {\centering\caption{\texttt{In-house}}}
  \vspace{0.95cm}
\end{subfigure}
\hfill
\begin{subfigure}[b]{0.16\textwidth}
    \includegraphics[height=2.1cm, width= 2 cm]{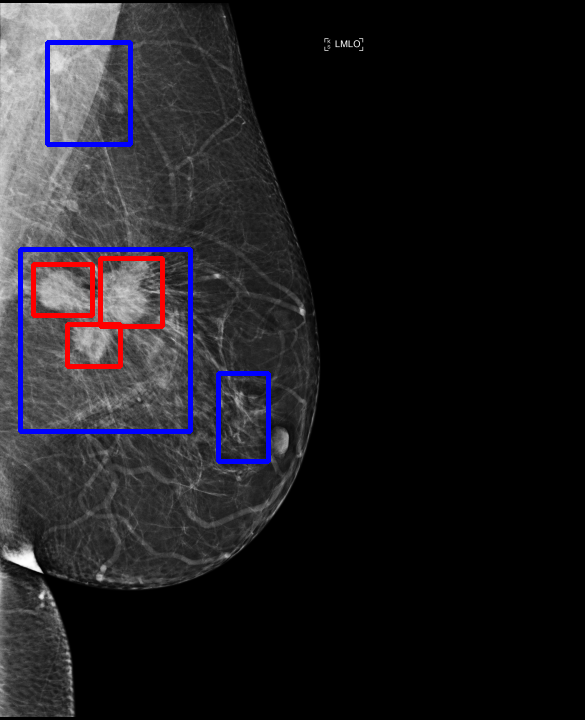}
\end{subfigure}
\hfill
\begin{subfigure}[b]{0.16\textwidth}
    \includegraphics[height=2.1cm, width= 2 cm]{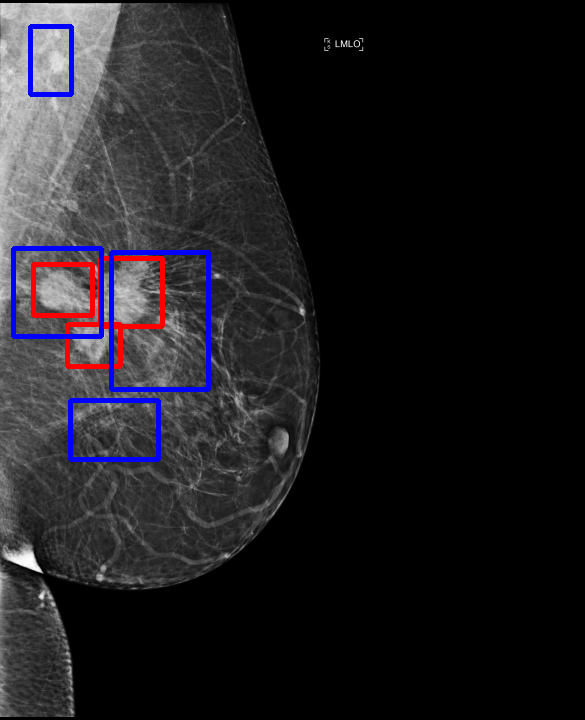}
\end{subfigure}
\hfill
\begin{subfigure}[b]{0.16\textwidth}
    \includegraphics[height=2.1cm, width= 2 cm]{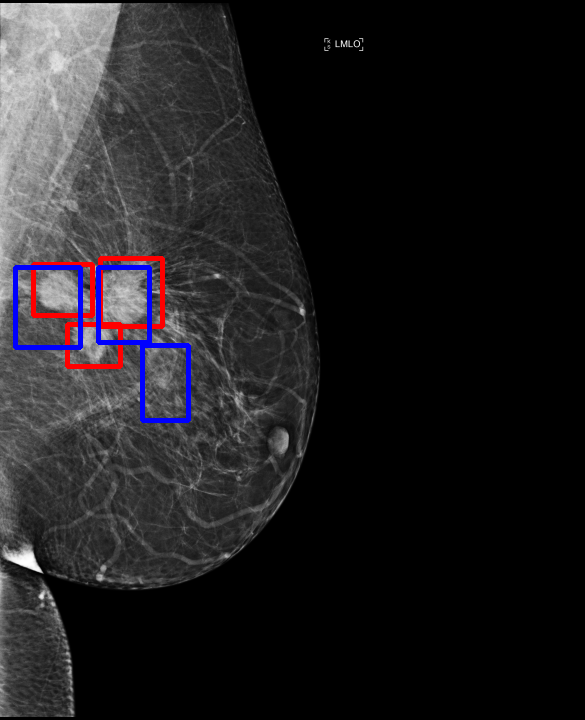}
\end{subfigure}
\hfill
\begin{subfigure}[b]{0.16\textwidth}
    \includegraphics[height=2.1cm, width= 2 cm]{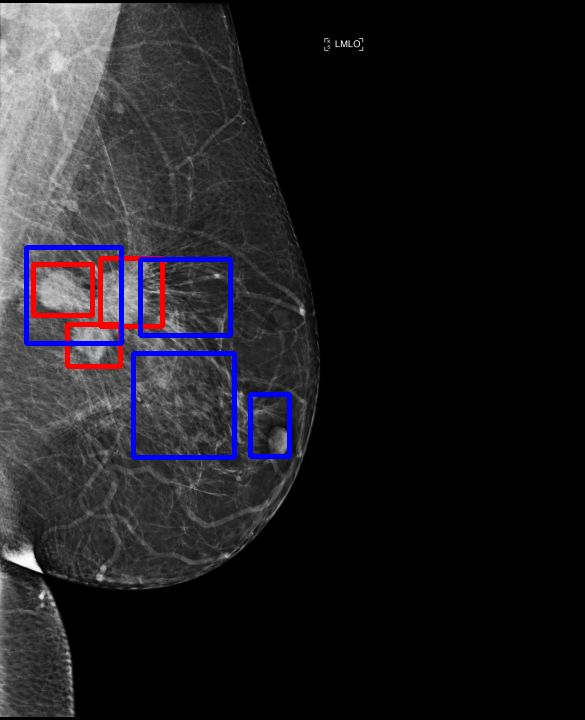}
\end{subfigure}
\hfill
\begin{subfigure}[b]{0.16\textwidth}
    \includegraphics[height=2.1cm, width= 2 cm]{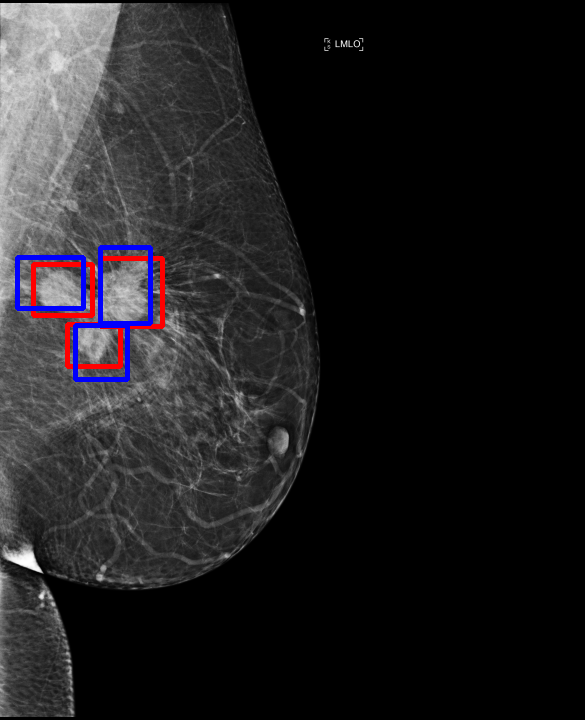}
\end{subfigure}
\vspace{-0.02em}
\begin{subfigure}[b]{0.15\textwidth}
  {\centering\caption{\ddsm~\cite{lee2017curated}}}
  \vspace{0.95cm}
\end{subfigure}
\hfill
\begin{subfigure}[b]{0.16\textwidth}
    \includegraphics[height=2.1cm, width= 2 cm]{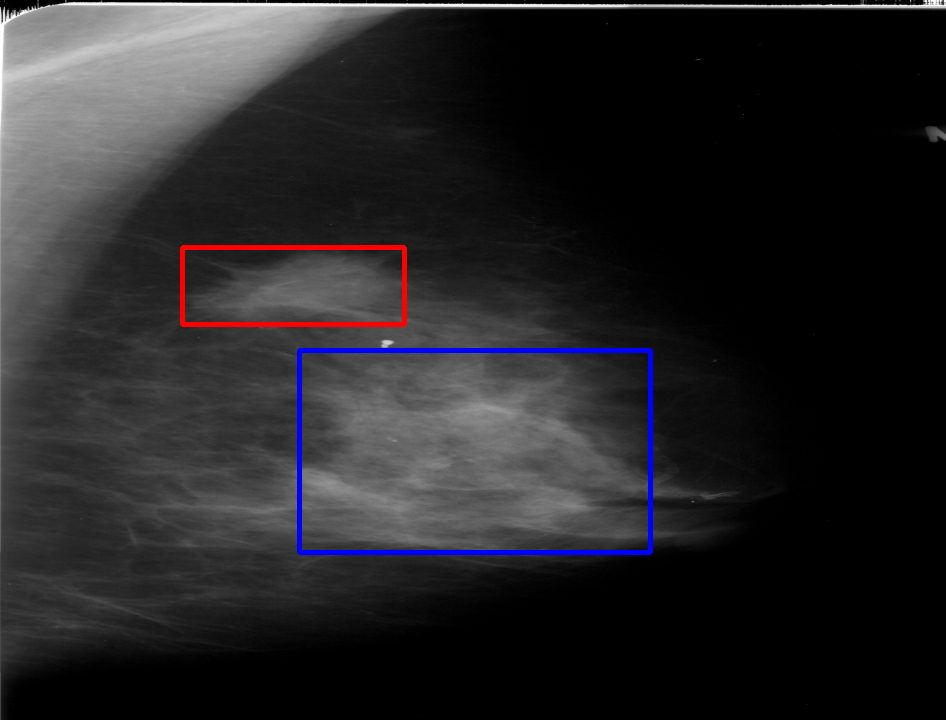}
\end{subfigure}
\hfill
\begin{subfigure}[b]{0.16\textwidth}
    \includegraphics[height=2.1cm, width= 2 cm]{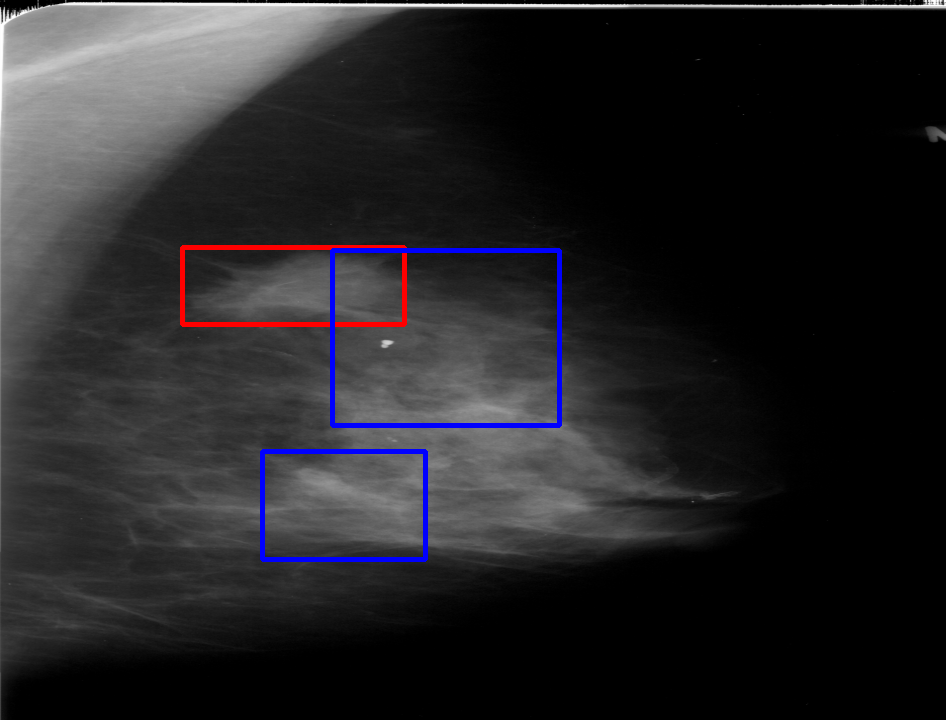}
\end{subfigure}
\hfill
\begin{subfigure}[b]{0.16\textwidth}
    \includegraphics[height=2.1cm, width= 2 cm]{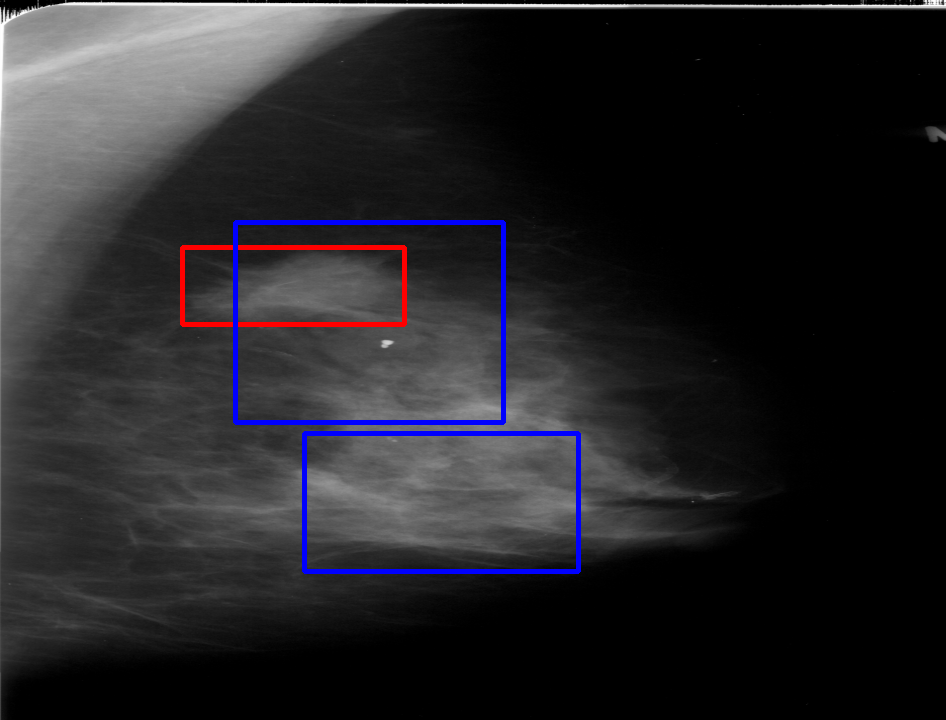}
\end{subfigure}
\hfill
\begin{subfigure}[b]{0.16\textwidth}
    \includegraphics[height=2.1cm, width= 2 cm]{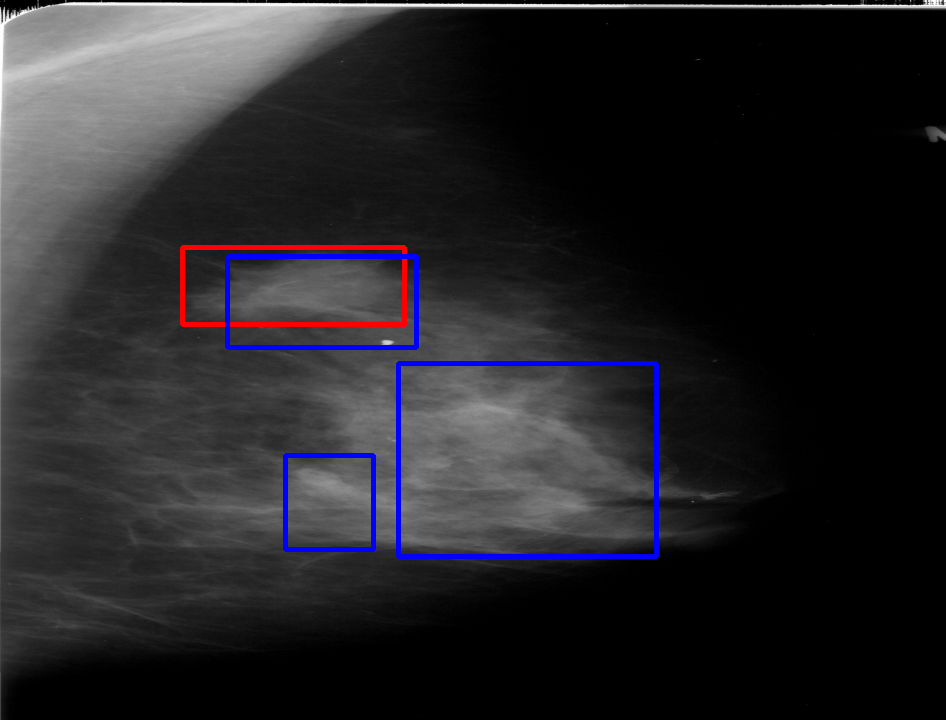}
\end{subfigure}
\hfill
\begin{subfigure}[b]{0.16\textwidth}
    \includegraphics[height=2.1cm, width= 2 cm]{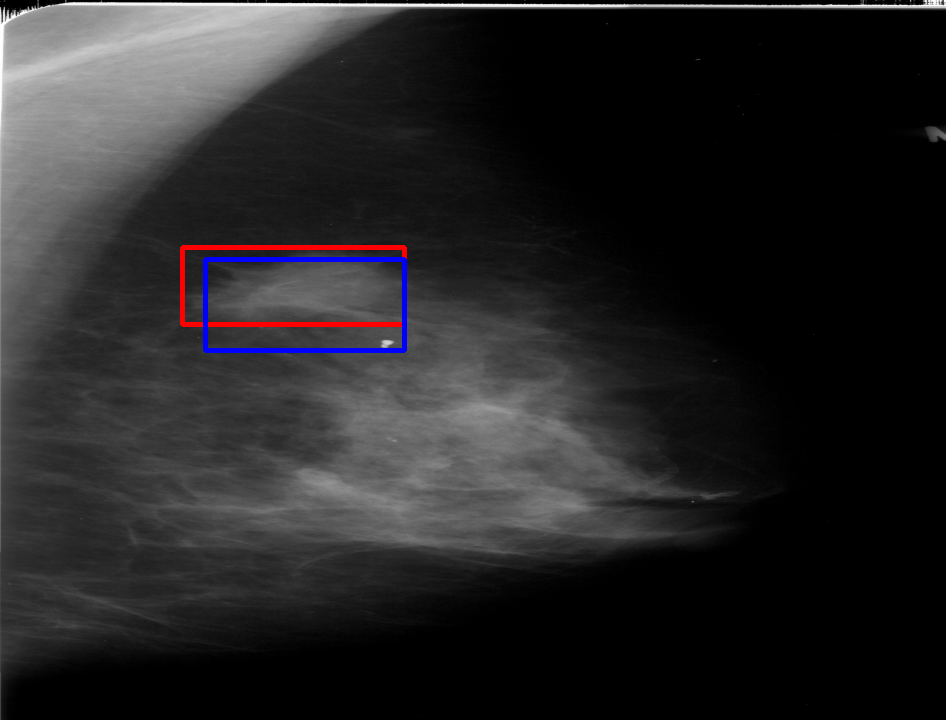}
\end{subfigure}
\vspace{-0.01em}
\begin{subfigure}[b]{0.15\textwidth}
  {\centering\caption{\rsna}}
  \vspace{1.55cm}
\end{subfigure}
\hfill
\begin{subfigure}[b]{0.16\textwidth}
    \includegraphics[height=2.1cm, width=2cm]{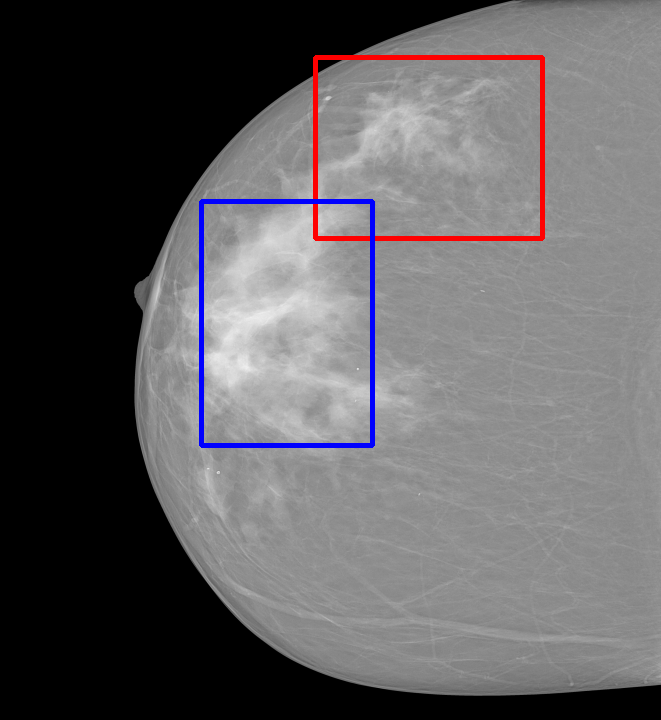}
    \caption{Source}
\end{subfigure}
\hfill
\begin{subfigure}[b]{0.16\textwidth}
    \includegraphics[height=2.1cm, width=2cm]{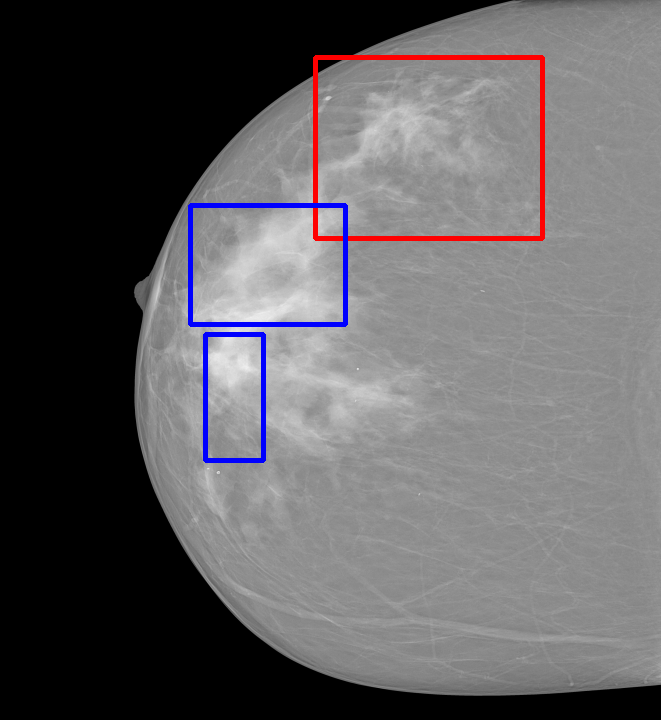}
    \caption{AT~\cite{li2022cross}}
\end{subfigure}
\hfill
\begin{subfigure}[b]{0.16\textwidth}
    \includegraphics[height=2.1cm, width=2cm]{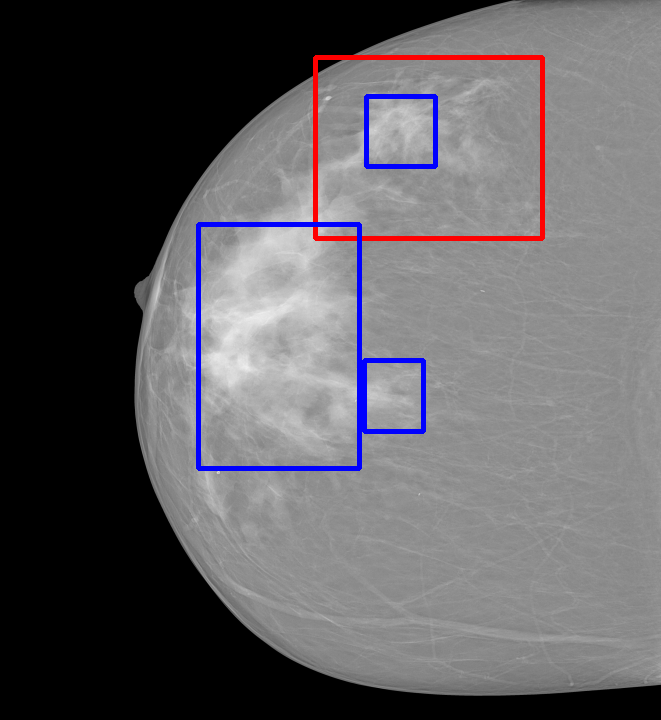}
    \caption{MRT~\cite{zhao2023masked}}
\end{subfigure}
\hfill
\begin{subfigure}[b]{0.16\textwidth}
    \includegraphics[height=2.1cm, width=2cm]{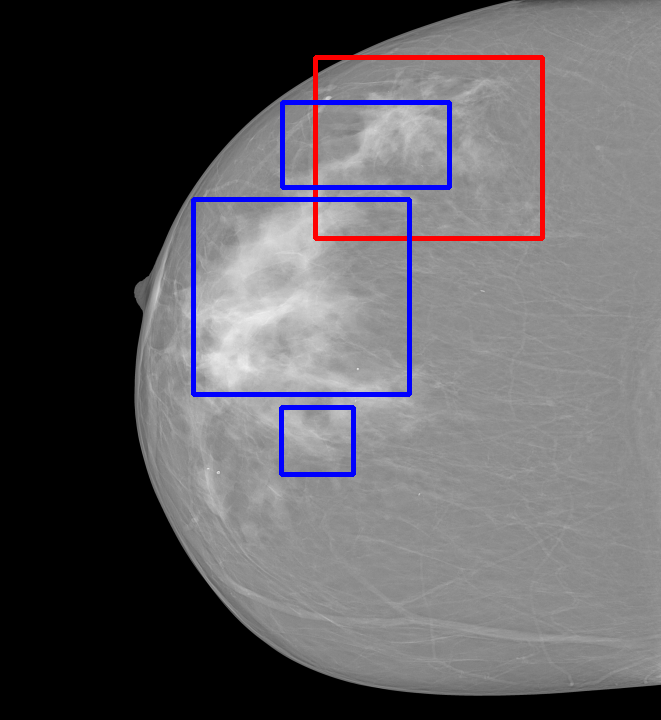}
    \caption{HT~\cite{deng2023harmonious}}
\end{subfigure}
\hfill
\begin{subfigure}[b]{0.16\textwidth}
    \includegraphics[height=2.1cm, width=2cm]{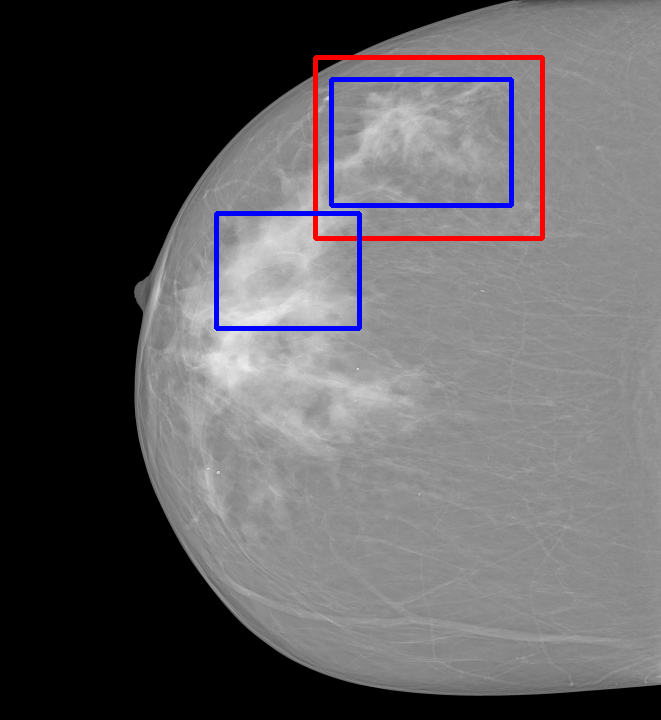}
    \caption{Ours}
\end{subfigure}
\caption{Qualitative result comparison on in-house, \ddsm, and \rsna datasets. Red boxes show the ground truth, and blue boxes show the predictions.}
\label{fig:qualitative_results}
\end{figure}